\pdfoutput=1

\documentclass[11pt]{article}

\usepackage{acl}

\usepackage{times}
\usepackage{latexsym}
\usepackage{graphicx}
\usepackage{caption}
\usepackage{hyperref}
\usepackage{subcaption}
\usepackage{booktabs}
\usepackage{microtype}
\usepackage{makecell}

\usepackage{xcolor}
\usepackage[export]{adjustbox}

\usepackage[T1]{fontenc}

\usepackage[utf8]{inputenc}

\usepackage{microtype}

\usepackage{amssymb}
\usepackage{utfsym}

\newcommand{\greencheck}{{\color{green}\usym{2714}}}
\newcommand{\redcross}{{\color{red}\usym{2718}}}

\title{IFAN: An Explainability-Focused Interaction Framework \\ for Humans and NLP Models}

\author{Edoardo Mosca, Daryna Dementieva, Tohid Ebrahim Ajdari, \\
{\bf Maximilian Kummeth}, {\bf{Kirill Gringauz}}, {\bf Yutong Zhou} \and {\bf Georg Groh} \\
TU Munich, Department of Informatics, Germany\\ 
\texttt{\{name.surname\}@tum.de} \\
\texttt{grohg@in.tum.de}}

\begin{document}
\maketitle
\begin{abstract}
Interpretability and human oversight are fundamental pillars of deploying complex NLP models into real-world applications. However, applying explainability and human-in-the-loop methods requires technical proficiency. Despite existing toolkits for model understanding and analysis, options to integrate human feedback are still limited. We propose IFAN, a framework for real-time explanation-based interaction with NLP models. Through IFAN's interface, users can provide feedback to selected model explanations, which is then integrated through adapter layers to align the model with human rationale. We show the system to be effective in debiasing a hate speech classifier with minimal impact on performance. IFAN also offers a visual admin system and API to manage models (and datasets) as well as control access rights. A demo is live at \href{https://ifan.ml/}{ifan.ml}.
\end{abstract}

\section{Introduction}

As \emph{Natural Language Processing} (NLP) systems continue to improve in performance, they are increasingly adopted in real-world applications \citep{khurana2022natural}. \emph{Large Language Models} (LLMs)---such as GPT-3 \citep{brown2020language}, BLOOM \citep{scao2022bloom}, and T5 \citep{raffel2020t5}---are without a shred of doubt the main protagonists of recent advances in the field. They are able to substantially outperform previous solutions while being directly applicable to any NLP task.

\begin{figure}
    \centering
    \includegraphics[width=0.95\linewidth]{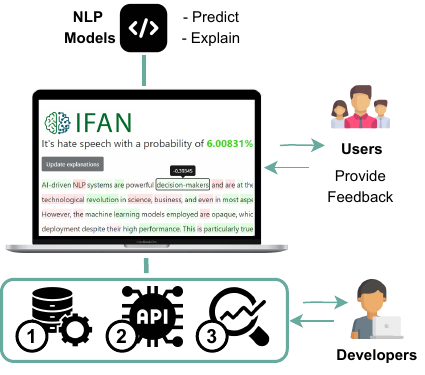}
    \caption{IFAN in brief. The interface allows NLP models and users to interact through predictions, explanations, and feedback. IFAN also provides developers with (1) a manager for models and datasets, (2) model API access, and (3) reports about the model.}
    \label{fig:ifan_intro}
\end{figure}

There are however strong concerns given the black-box nature of such architectures \citep{madsen2022posthoc, mosca-etal-2022-suspicious}. In fact, their large scale and high complexity are substantial drawbacks in terms of \emph{transparency}, \emph{accountability}, and \emph{human oversight}. Beyond ethical considerations, even legal guidelines from the European Union are now explicitly defining these interpretability factors as essential for any deployed AI system \citep{european2020whitepaper}.

Research efforts in \emph{eXplainable Artificial Intelligence} (XAI) \citep{arrieta2020explainable, mosca-etal-2022-shap} and \emph{Human-in-the-Loop} (HitL) machine learning \citep{monarch2021human} have thus been on the rise---producing solutions that aim at mitigating the current lack of interpretability. Most notably, the recent literature contains a number of toolkits and frameworks to analyze, understand, and improve complex NLP models \citep{wallace-etal-2019-allennlp, liu-etal-2021-explainaboard}. Some of them even offer low-code interfaces for stakeholders who do not possess the otherwise required technical proficiency. Nonetheless, current options to collect human rationale and provide it as feedback to the model are still limited.

We propose IFAN, a novel low-to-no-code framework to interact in real time with NLP models via explanations. Our contribution can be summarized as follows:

\begin{itemize}
    \item[\textbf{(1)}] IFAN offers an interface for users to provide feedback to selected model explanations, which is then integrated via parameter-efficient adapter layers.
    \item[\textbf{(2)}] Our live platform also offers a visual administration system and API to manage models, datasets, and users as well as their corresponding access rights.
    \item[\textbf{(3)}] We show the efficiency of our framework in debiasing a hate speech classifier and propose a feedback-rebalancing step to mitigate the model's forgetfulness across updates.
\end{itemize}

IFAN's demo is accessible at ifan.ml\footnote{\href{https://ifan.ml}{https://ifan.ml}}  together with its documentation.\footnote{\href{https://ifan.ml/documentation}{https://ifan.ml/documentation}} 
Full access is available with login credentials, which we can provide upon request. A supplementary video showcase can be found online\footnote{\href{https://youtu.be/EzC6HI3JwaQ}{https://youtu.be/EzC6HI3JwaQ}}.

\section{Related Work}

\subsection{HitL with Model Explanations}

\emph{Human-in-the-Loop} (HitL) machine learning studies how models can be continuously improved with human feedback \citep{monarch2021human}. While a large part of the HitL literature deals with label-focused feedback such as \emph{active learning}, more recent works explore how explanations can be leveraged to provide more detailed human rationale \citep{lertvittayakumjorn-toni-2021-explanation}.

Combining classical HitL \citep{wang-etal-2021-putting} with explanations to construct human feedback for the model \citep{han-etal-2020-explaining} has been referred to as \emph{Explanation-Based Human Debugging} (EBHD) \citep{lertvittayakumjorn-toni-2021-explanation}. Good examples are \citet{ray2019can}, \citet{selvaraju2019taking}, and \citet{strout-etal-2019-human}, which show improvements in performance and interpretability when iteratively providing models with human rationale.

A more NLP-focused EBHD approach is \citet{yao2021refining}, where the authors leverage explanations to debug and refine two transformer instances---BERT \citep{devlin-etal-2019-bert} and RoBERTa \citep{liu2019roberta}. Concretely, word saliency explanations at different levels of granularity are provided to humans, who in turn provide suggestions in the form of natural language. The annotator's feedback is converted into first-order logic rules, which are later utilized to condition learning with new samples. 

\subsection{Interactive NLP Analysis Platforms}

In the recent literature, we can find strong contributions in terms of software and digital toolkits to analyze and explain NLP models \cite{wallace-etal-2019-allennlp, hoover-etal-2020-exbert} as well as further refining them via parameter-efficient fine-tuning \citep{beck-etal-2022-adapterhub}. 

For instance, \citet{liu-etal-2021-explainaboard} proposes \textsc{ExplainaBoard}, an interactive explainability-focused leaderboard for NLP models. More in detail, it allows researchers to run diagnostics about the strengths and weaknesses of a given model, compare different architectures, and closely analyze predictions as well as recurring model mistakes. Similarly, the \textsc{Language Interpretability Tool} by \citet{tenney-etal-2020-language} is an open-source platform and API to visualize and understand NLP models. In particular, it provides a browser-based interface integrating local explanations as well as counterfactual examples to enable model interpretability and error analysis. 

Finally, \citet{beck-etal-2022-adapterhub} releases \textsc{AdapterHub Playground}, a no-code platform to few-shot learning with language models. Specifically, the authors built an intuitive interface where users can easily perform predictions and training of complex NLP models on several natural language tasks.

\begin{figure*}[th!]
    \centering
    \includegraphics[width=0.9\textwidth]{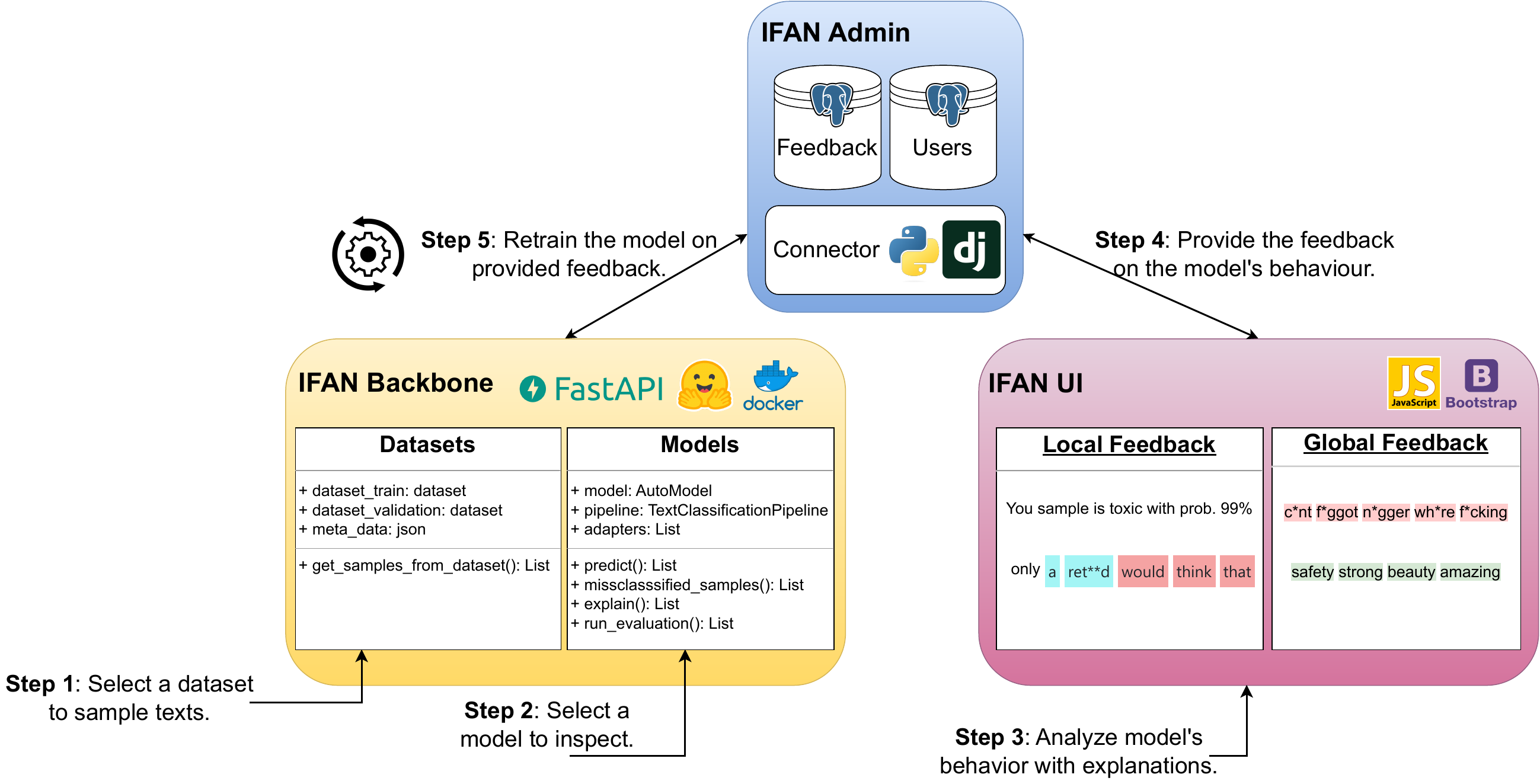}
    \caption{Overall schema of IFAN idea: (i)~The user selects a dataset or writes a customized input. (ii)~Then the user can select a model which should be inspected. (iii)~With the UI, annotators can check the model's prediction on a sample and two types of explanations -- local and global. (iv)~If there is some misbehavior, the annotators can provide feedback. (v)~The feedback is stored and then used to fine-tune the model.}
    \label{fig:ifan}
\end{figure*}

\section{IFAN}
The \textbf{I}nteraction \textbf{F}ramework for \textbf{A}rtificial and \textbf{N}atural Intelligence (\textbf{IFAN}) is a web-based platform for inspecting and controlling text processing models. Its main goal is to decrease the opacity of NLP systems and integrate explanation-based HitL into their development pipeline. Through our interface, stakeholders can test and explain models' behavior and---when encountering anomalies in predictions or explanations---they can fix them onsite by providing feedback.

The main blocks of the platform are presented in Figure~\ref{fig:ifan}. The \textbf{Backbone} part contains all machine learning development components---datasets and models. We adopt HuggingFace formats (see \ref{subsec:datasets} and \ref{subsec:models}) \cite{wolf-etal-2020-transformers} and wrap the entire backbone as a Docker\footnote{\href{https://www.docker.com}{https://www.docker.com}} image for deployment. The \textbf{User Interface} is the visual component of the platform, where all the human-machine interaction takes place. Here, developers have also access to additional visual resources to configure details about models, datasets, and users.

The connection between the backbone and the user interface is managed by the \textbf{Admin} component. All the user data and rights as well as samples receiving feedback are stored in a PostgreSQL\footnote{\href{https://www.postgresql.org}{https://www.postgresql.org}} database instance. The communication is handled via Python Django\footnote{\href{https://www.djangoproject.com}{https://www.djangoproject.com}}, which integrates everything w.r.t. user authentication, API calls/responses, state logs, and location of backbone resources. In the next sections, we provide a more detailed description of the main platform components.

\subsection{User Interface} \label{subsec:user_inteface}

Our frontend is built with Bootstrap\footnote{\href{https://getbootstrap.com}{https://getbootstrap.com}} and JavaScript\footnote{\href{https://www.javascript.com}{https://www.javascript.com}}. Currently, the pages available in our UI are the following:

\paragraph{Landing Page} Here users can get a short introduction to IFAN. We briefly explain our platform's goals, the concept of HitL, and how our framework can be integrated into the development of NLP models.

\paragraph{Documentation} It provides a detailed description of all the UI components together with screenshots and guidelines. Here, users can find specific instructions on how to configure and interact with our platform.

\paragraph{Feedback} This is the main interaction page. Here, users can run a model on an input sample either taken from the dataset or that they wrote themselves. Then, they can load the model's prediction and explanations and provide feedback in terms of both the label and features' relevance.

\paragraph{Report} This page has limited access (see \ref{subsec:users}). Developers can evaluate models before and after feedback incorporation on a chosen dataset as well as inspect misclassified samples.

\paragraph{Configuration} This page has limited access (see \ref{subsec:users}). Here, developers can configure and manage the platform, More specifically, users can be created, modified, and deleted as well as upgraded or downgraded in their roles and access rights. Also, they can manage models and datasets as well as specify the currently active ones.

\paragraph{Account Settings} Each authorized user can view, edit, export, and delete their account data (GDPR compliance) as well as reset their login password.

\subsection{Users} \label{subsec:users}

The platform separates users in three tiers: \textit{developers}, \textit{annotators}, and \textit{unauthorized} users (Table~\ref{tab:users}). 

\begin{table}[h!]
    \centering
    \footnotesize
    \begin{tabular}{c|c|c|c}
    \toprule
         & Dev & Annotator & Unauthorized \\
        \hline
        \makecell{Classification \\ \& Explanations} & \greencheck & \greencheck & \greencheck \\
        \hline
        \makecell{Smart Samples \\ Selection} & \greencheck & \greencheck & \redcross \\
        \hline
        Feedback & \greencheck & \greencheck & \redcross \\
        \hline
        \makecell{Active \\ Configuration} & \greencheck & \redcross & \redcross \\
        \hline
        \makecell{Model Report \&\\Miscl. Samples} & \greencheck & \redcross & \redcross \\
        \hline
        \makecell{New Models \& \\ Datasets Upload} & \greencheck & \redcross & \redcross \\
        \hline
        \makecell{New Users \\ Creation} & \greencheck & \redcross & \redcross \\
    \bottomrule
    \end{tabular}
    \caption{Different levels of access to IFAN.}
    \label{tab:users}
\end{table}

Unauthorized users do not possess login credentials and have limited access to the platform. They can visualize model predictions and explanations but their feedback is not considered.

Normal users (or annotators) are known through their credentials and can thus actively engage with the model. During a HitL iteration, they can use the feedback page with pre-configured datasets and models, test the model on a text sample, view explanations, and provide feedback if needed. 

Developers have full access and can configure all aspects of the platform. More specifically, they have access to the \emph{report} and \emph{configuration} pages (see \ref{subsec:user_inteface}) and can thus manage anything regarding users, roles, API access, models, and datasets. 

\subsection{Datasets} \label{subsec:datasets}

Before the model's behavior exploration, the \emph{active dataset} should be specified via the configuration page (see \ref{subsec:user_inteface}). This is the dataset from which the text examples for the model testing are sampled.

\begin{table}[h!]
    \centering
    \footnotesize
    \begin{tabular}{p{3cm}|p{4cm}}
    \toprule
        \textbf{Dataset} & \textbf{Short Description} \\
        \hline
        HateXplain \cite{DBLP:conf/aaai/MathewSYBG021} & A dataset for hate speech classification which has 3 classes for hate type detection, the target community classification, and rationales. \\
        \hline
        GYAFC \cite{rao-tetreault-2018-dear} & Formality detection dataset which corresponds to 2-class classification: formal and informal. \\
    \bottomrule
    \end{tabular}
    \caption{Example of datasets tested at IFAN.}
    \label{tab:datasets}
\end{table}

We conform to a standard format by using the HuggingFace Datasets library\footnote{\href{https://huggingface.co/docs/datasets/index}{https://huggingface.co/docs/datasets/index}}. Developers interacting with our platform are strongly encouraged to adhere to this standard when uploading new datasets and making them available to the interface. Table~\ref{tab:datasets} shows two examples of datasets already available on our platform.

\subsection{Models} \label{subsec:models}

Analogous to datasets, the \emph{active model} should be specified via the configuration page and should adhere to the HuggingFace Models standard.\footnote{\href{https://huggingface.co/docs/transformers/main\_classes/model}{https://huggingface.co/docs/transformers/main\_classes/model}}

\begin{figure}[h!]
    \centering
    \includegraphics[width=0.48\textwidth]{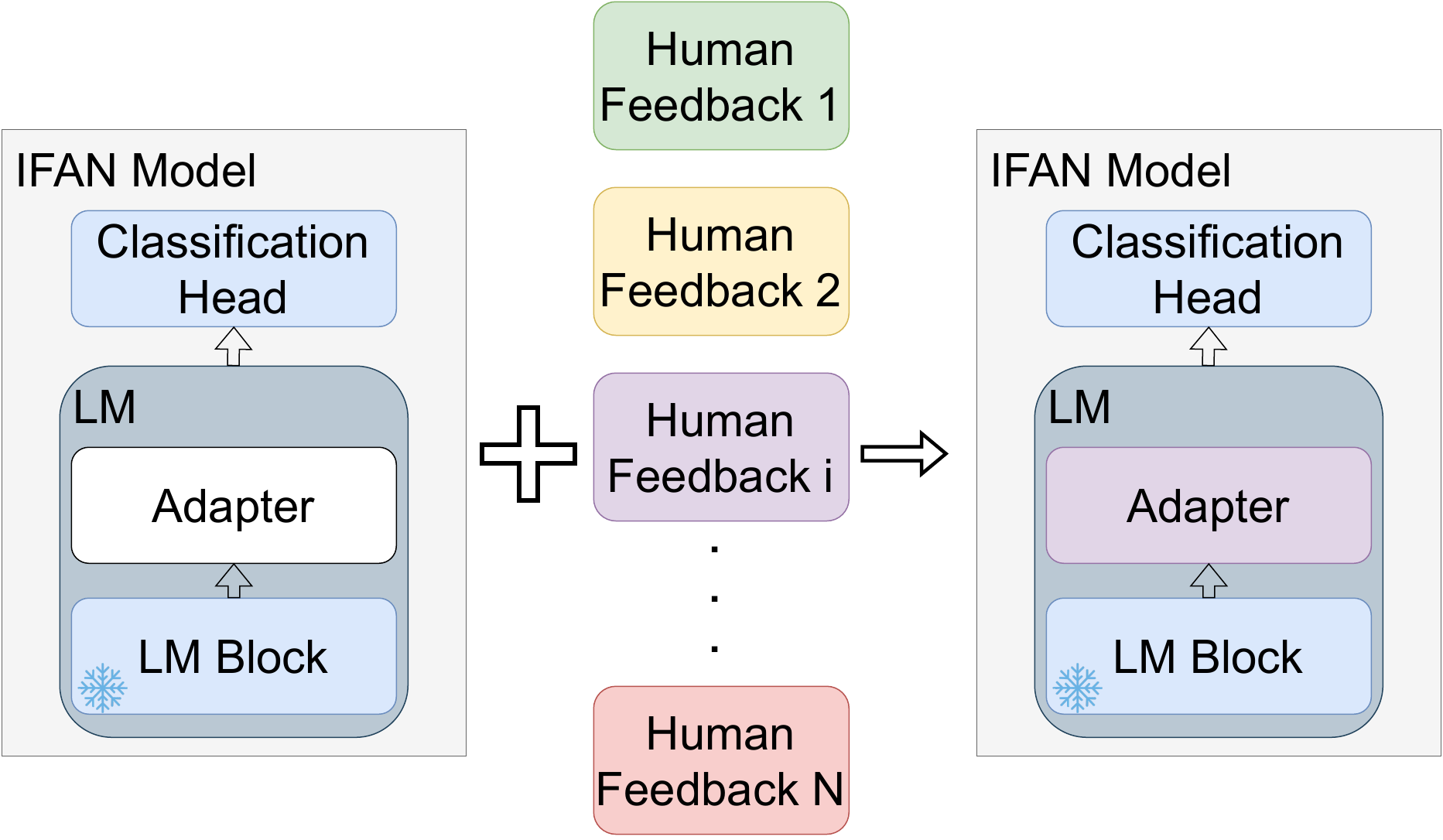}
    \caption{The proposed architecture for the models integrated into IFAN: addition of Adapter layer which is trainable on provided human feedback.}
    \label{fig:ifan_model}
\end{figure}

To incorporate feedback into our models, we utilize adapter layers \cite{DBLP:conf/icml/HoulsbyGJMLGAG19}, a parameter-efficient fine-tuning technique. Figure~\ref{fig:ifan_model} sketches an overview of the architecture used. Adapters are integrated on top of each language model unit (e.g. transformer block) and are trained with the human feedback while we freeze all other model weights. Adapters can also be disabled to recover to the original state of the model.


\subsection{Explanations \& Feedback Mechanism}

Users can evaluate the active model on the active dataset through the Feedback page. They may input text in three ways: i) create a text sample themselves; if authorized: ii) sample a random text from the active dataset; iii) sample a random \textit{misclassified} text from the test part of the active dataset. Users receive the classification results and the model's confidence. They can assess the result and correct any misclassifications. 

\begin{figure}[h!]
    \centering
    \includegraphics[width=0.45\textwidth]{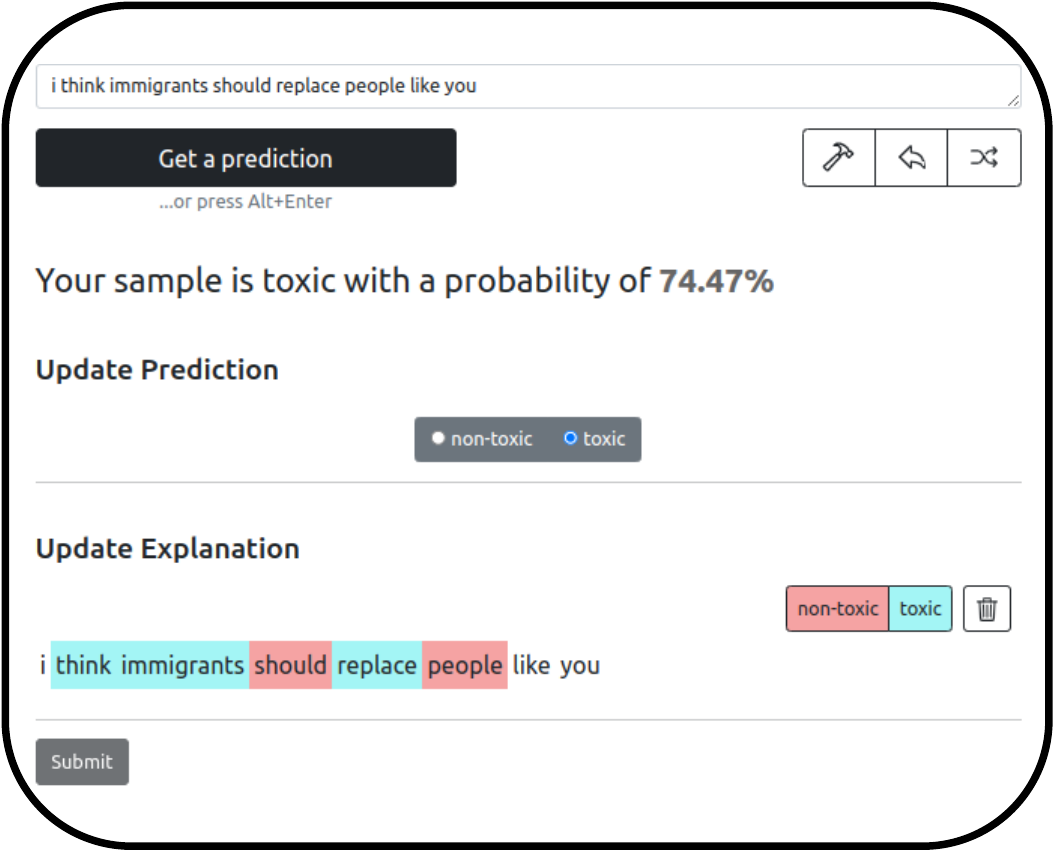}
    \caption{The example of the results and local explanations that annotators can obtain on the Feedback page.}
    \label{fig:explanation_example}
\end{figure}

To further inspect the model's behavior, we provide two types of explanations---local and global. For local explanations on a text sample, we display relevant features to each output class (Figure~\ref{fig:explanation_example}). We attribute scores using the LIME framework \citep{ribeiro2016should} and---to filter weak correlations---we highlight as relevant only tokens with a score above the threshold $\theta=0.1$. On the global side, we list the most influential unigrams for each output class. These can be inspected to extract insights about what keywords and patterns the model focuses on at the dataset level. For all 1-grams present in a dataset, their corresponding classification scores are calculated and the tokens with top scores are displayed on the page.

Annotators can easily edit the highlighted tokens and send the updated explanation as feedback. We store the result---i.e. the highlighted relevant parts---and use them to fine-tune the adapter layers. Freezing all other model weights minimizes the computational effort of the feedback step.

Regarding the fine-tuning procedure, directly using the highlighted feedback text for adapter fine-tuning causes significant losses in the original model performance. We propose to mix feedback with original samples to mitigate this effect, which allows effective feedback incorporation while reducing model forgetfulness (see \ref{sec:case_study} for more details).

\subsection{Backbone API}

We expose our backbone's API
to make available all essential dataset/model management functions. These provide a high-level interface for additional experiments dealing with model evaluation, explanation, and feedback. The API is built with the Python framework FastAPI\footnote{\href{https://fastapi.tiangolo.com}{https://fastapi.tiangolo.com}}, more details can be found in the Appendix \ref{sec:appendix_backbone_api}.

\section{Case Study} \label{sec:case_study}

We carried out a case study to test the applicability of IFAN. We chose a hate speech detection task based on the HateXplain dataset \cite{DBLP:conf/aaai/MathewSYBG021}. The goal of the experiment was to use our framework to debias a given hate speech detector.

Firstly, we modified the original dataset for binary classification task---\textit{``toxic''} and \textit{``non-toxic''}. We choose the Jewish subgroup as a target for our debiasing process. We fine-tuned BERT \cite{devlin-etal-2019-bert}\footnote{\href{https://huggingface.co/google/bert\_uncased\_L-2\_H-128\_A-2}{https://huggingface.co/google/bert\_uncased\_L-2\_H-128\_A-2}} and gave feedback to it. Additional experiments with BLOOM and other LLMs are provided in Appendix~\ref{sec:app_bloom} and \ref{sec:app_llm} respectively.

We annotate $24$ random misclassified samples, $12$ with the most confidence and $12$ with the least confidence scores (see Appendix~\ref{sec:app_samples}). We invited $3$ annotators to participate in the annotation process.
%
The n-grams that were modified by annotators were saved and used to create a new training dataset for the adapters. As a result, we collected $40$ annotated n-grams and repeated them to get $120$ training samples. To complete the new training creation, we balanced these samples with $500$ original samples ($250$ toxic, $250$ non-toxic) randomly selected from the HateXplain dataset.

\begin{table}[h!]
    \centering
    \footnotesize
    \begin{tabular}{p{3.5cm}|c|c|c|c}
        \toprule
        \textbf{Model} & \textbf{Pr} & \textbf{Re} & \textbf{F1} & \textbf{Pr$_{J}$} \\
        \hline
        BERT (baseline) & 0.80 & \textbf{0.78} & \textbf{0.79} & 0.95 \\
        \hline
        \multicolumn{5}{c}{\textit{Most Confident Missclassified}} \\ 
        \hline
        BERT+Feedback (non-bal.) & 0.34 & 0.28 & 0.31 & 0.82 \\
        BERT+Feedback (bal.) &  0.78 & 0.80 & \textbf{0.79} & \textbf{0.97} \\
        \hline
        \multicolumn{5}{c}{\textit{Least Confident Missclassified}} \\ 
        \hline
        BERT+Feedback (non-bal.) & \textbf{0.83} & 0.73 & 0.78 & 0.96\\
        BERT+Feedback (bal.) & 0.79 & \textbf{0.78} & 0.78 & 0.96\\
        \bottomrule
    \end{tabular}
    \caption{The results of the case study: hate speech classification model debiasing. We compare different strategies for feedback incorporation. Pr$_{J}$ states for the Precision score on the Jewish target group.}
    \label{tab:case_study_results}
\end{table}

The results are presented in Table~\ref{tab:case_study_results}. 
We observe that the non-balanced training dataset, which only contains feedback on the most confidently misclassified samples, resulted in a significant decrease in performance. While the inclusion of feedback on least confident samples caused a slight decline in the overall F1 score, Adapter training on the balanced feedback led to an improvement in the precision score for the Jewish target group.

\begin{figure*}[th!]
    \centering
    \begin{subfigure}{0.46\textwidth}
        \includegraphics[width=0.97\textwidth]{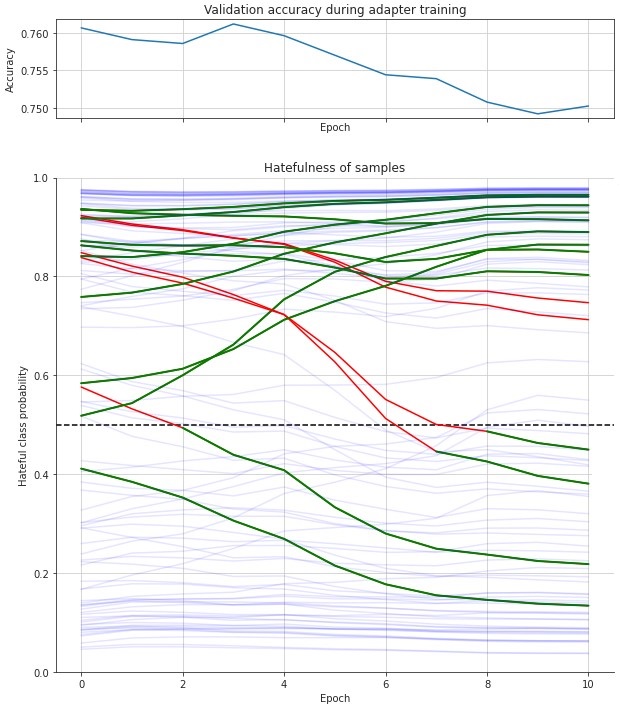}
        \caption{Training on feedback on the Jewish subgroup samples.}
    \end{subfigure}
    \begin{subfigure}{0.46\textwidth}
        \includegraphics[width=0.97\textwidth]{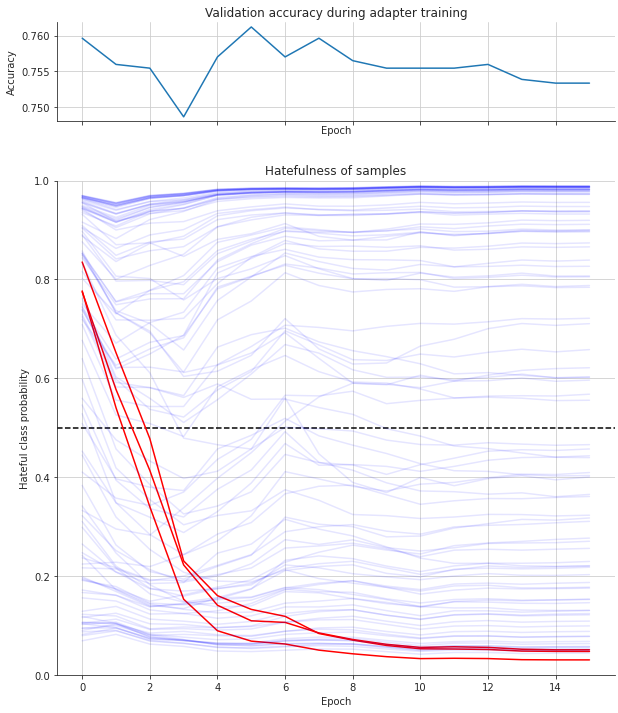}
        \caption{Training on feedback samples with ``jewish'' key-words.}
    \end{subfigure}
    \caption{Samples confidence variation as the model is fine-tuned with human feedback. The results of the domain case using IFAN platform. We can observe that for both experiments with balanced training data, the overall model's performance is only slightly changed while the model's behavior on the Jewish target group is improved.}
    \label{fig:case_study_results}
\end{figure*}
 
Figure \ref{fig:case_study_results} shows the changes in the detector while fine-tuning with the collected feedback. When rebalancing the feedback, only modified samples are drastically changed while the performance on the original texts is only slightly affected. A detailed comparison between fine-tuning on non-balanced and balanced feedback is in Appendix~\ref{sec:app_eddback_comparison}).

\section{Limitations \& Future Work} \label{sec:limitations}

As of now, our feedback system is limited to applications in the sequence-to-class format. Work on extending the platform to further task through LLM prompting is currently in progress (see Appendix~\ref{sec:app_llm}).

At the same time, we currently offer a limited set of explanation, feedback, and management options, which we plan to increase in the immediate future. A small user study has been conducted (Appendix \ref{sec:user_study}) to collect feedback about the platform and improve its user-friendliness. Our intent is to continue iterating the development of new features with trials with developers and laymen.

Finally, our experiments do not yet show clear trends w.r.t. the correlation between performance and feedback hyperparameters. Indeed, further research and trials have to be carried out to establish optimal choices for the number of feedback samples, fine-tuning epochs, and the rebalancing ratio.

\section{Conclusion}

This work proposes IFAN, a framework focusing on real-time explanation-based interaction between NLP models and human annotators. Our contribution is motivated by the limited options in terms of existing tools to interpret and control NLP models. 

IFAN is composed of three main units. The \textbf{Backbone} unifies all the machine learning pipelines and exposes an API for accessibility. The \textbf{User Interface}---organized in \emph{landing page}, \emph{documentation}, \emph{feedback}, \emph{report}, and \emph{configuration}---provides an intuitive visual component to interact with models. Finally, the \textbf{Admin} controls the connection between the two previous components.

Additionally, we introduce the feedback mechanism that takes advantage of adapter layers to efficiently and iteratively fine-tune models on the downstream task. Our experiments show the frameworks’ effectiveness at debiasing a hate speech classifier with minimal performance loss.


We believe IFAN to be a valuable step towards enabling the interpretable and controllable deployment of NLP models---allowing users with no technical proficiency to interact and provide feedback to deployed NLP systems. Regarding future work, we set as a priority to extend the framework to more NLP tasks as well as to integrate additional model analysis features and feedback mechanisms.


\section*{Acknowledgments}
We thank Natália Souza Soares and Ashish Jha for their valuable contribution. This paper has been supported by the \emph{German Federal Ministry of Education and Research} (BMBF, grant 01IS17049). 

\section*{Ethical Considerations}
In this work, we showed the experiments of hate speech model debasing. The hate speech detection task is the task that requires a lot of attention to provide a fair outcome. One of the issues still is bias, especially against minority groups due to prejudices. We aimed to show an example of how the model can be debiased with respect to some target racial groups. With a conscientious selection of annotators and feedback, we hope that our proposed platform will serve to efficiently adjust NLP models to the diverse world.

For these reasons, we also believe that interpretability and controllability of modern NLP models and systems are fundamental pillars for their ethical and safe deployment \citep{european2020whitepaper}. This works aims at having a positive impact on both aspects as it provides a tool to explain models and provide them with feedback. By reducing the technical proficiency required to interact with NLP systems, we hope to facilitate the process of providing valuable human rationales to influence complex models. 

Ensuring high quality for the human feedback is challenging \citep{al-kuwatly-etal-2020-identifying}, and exposing models to external influence can be used as an exploit by adversarial agents \citep{mosca-etal-2022-suspicious}. Especially with a very small crowd of annotators, there's potential for a few people to have a strong influence on the model. A restrictive access rights management system like IFAN's already mitigates these issues. We believe that additional security features as well as tracking annotators' impact are key for future work to foster their trustworthiness.

Previous works mention that users can feel discouraged and frustrated when interacting with poor models and badly-designed interfaces, which can also affect feedback quality \citep{lertvittayakumjorn-toni-2021-explanation}. This can be addressed by integrating user studies in the development process in order to design more intuitive interfaces and improve the overall user experience.

On the opposite end of the spectrum, plausible explanations can make humans overestimate the model's capabilities and make them trust systems that are still not ready for deployment. In this case, a more diverse and complementary set of explanations for users \citep{madsen2022posthoc} as well as comprehensive model reports for developers are core goals to provide a more complete picture of the models to be deployed.

\bibliography{anthology,custom}

\begin{thebibliography}{31}
\expandafter\ifx\csname natexlab\endcsname\relax\def\natexlab#1{#1}\fi

\bibitem[{Al~Kuwatly et~al.(2020)Al~Kuwatly, Wich, and
  Groh}]{al-kuwatly-etal-2020-identifying}
Hala Al~Kuwatly, Maximilian Wich, and Georg Groh. 2020.
\newblock \href {https://doi.org/10.18653/v1/2020.alw-1.21} {Identifying and
  measuring annotator bias based on annotators{'} demographic characteristics}.
\newblock In \emph{Proceedings of the Fourth Workshop on Online Abuse and
  Harms}, pages 184--190, Online. Association for Computational Linguistics.

\bibitem[{Arrieta et~al.(2020)Arrieta, D{\'\i}az-Rodr{\'\i}guez, Del~Ser,
  Bennetot, Tabik, Barbado, Garc{\'\i}a, Gil-L{\'o}pez, Molina, Benjamins
  et~al.}]{arrieta2020explainable}
Alejandro~Barredo Arrieta, Natalia D{\'\i}az-Rodr{\'\i}guez, Javier Del~Ser,
  Adrien Bennetot, Siham Tabik, Alberto Barbado, Salvador Garc{\'\i}a, Sergio
  Gil-L{\'o}pez, Daniel Molina, Richard Benjamins, et~al. 2020.
\newblock Explainable artificial intelligence (xai): Concepts, taxonomies,
  opportunities and challenges toward responsible ai.
\newblock \emph{Information Fusion}, 58:82--115.

\bibitem[{Beck et~al.(2022)Beck, Bohlender, Viehmann, Hane, Adamson, Khuri,
  Brossmann, Pfeiffer, and Gurevych}]{beck-etal-2022-adapterhub}
Tilman Beck, Bela Bohlender, Christina Viehmann, Vincent Hane, Yanik Adamson,
  Jaber Khuri, Jonas Brossmann, Jonas Pfeiffer, and Iryna Gurevych. 2022.
\newblock \href {https://doi.org/10.18653/v1/2022.acl-demo.6} {{A}dapter{H}ub
  playground: Simple and flexible few-shot learning with adapters}.
\newblock In \emph{Proceedings of the 60th Annual Meeting of the Association
  for Computational Linguistics: System Demonstrations}, pages 61--75, Dublin,
  Ireland. Association for Computational Linguistics.

\bibitem[{Brown et~al.(2020)Brown, Mann, Ryder, Subbiah, Kaplan, Dhariwal,
  Neelakantan, Shyam, Sastry, Askell et~al.}]{brown2020language}
Tom Brown, Benjamin Mann, Nick Ryder, Melanie Subbiah, Jared~D Kaplan, Prafulla
  Dhariwal, Arvind Neelakantan, Pranav Shyam, Girish Sastry, Amanda Askell,
  et~al. 2020.
\newblock Language models are few-shot learners.
\newblock \emph{Advances in neural information processing systems},
  33:1877--1901.

\bibitem[{Devlin et~al.(2019)Devlin, Chang, Lee, and
  Toutanova}]{devlin-etal-2019-bert}
Jacob Devlin, Ming-Wei Chang, Kenton Lee, and Kristina Toutanova. 2019.
\newblock \href {https://doi.org/10.18653/v1/N19-1423} {{BERT}: Pre-training of
  deep bidirectional transformers for language understanding}.
\newblock In \emph{Proceedings of the 2019 Conference of the North {A}merican
  Chapter of the Association for Computational Linguistics: Human Language
  Technologies, Volume 1 (Long and Short Papers)}, pages 4171--4186,
  Minneapolis, Minnesota. Association for Computational Linguistics.

\bibitem[{{European Commission}(2020)}]{european2020whitepaper}
{European Commission}. 2020.
\newblock \href
  {https://commission.europa.eu/publications/white-paper-artificial-intelligence-european-approach-excellence-and-trust_en}
  {White paper on artificial intelligence: a european approach to excellence
  and trust}.
\newblock \emph{Com (2020) 65 Final}.

\bibitem[{Han et~al.(2020)Han, Wallace, and
  Tsvetkov}]{han-etal-2020-explaining}
Xiaochuang Han, Byron~C. Wallace, and Yulia Tsvetkov. 2020.
\newblock \href {https://doi.org/10.18653/v1/2020.acl-main.492} {Explaining
  black box predictions and unveiling data artifacts through influence
  functions}.
\newblock In \emph{Proceedings of the 58th Annual Meeting of the Association
  for Computational Linguistics}, pages 5553--5563, Online. Association for
  Computational Linguistics.

\bibitem[{Hoover et~al.(2020)Hoover, Strobelt, and
  Gehrmann}]{hoover-etal-2020-exbert}
Benjamin Hoover, Hendrik Strobelt, and Sebastian Gehrmann. 2020.
\newblock \href {https://doi.org/10.18653/v1/2020.acl-demos.22} {ex{BERT}: {A}
  {V}isual {A}nalysis {T}ool to {E}xplore {L}earned {R}epresentations in
  {T}ransformer {M}odels}.
\newblock In \emph{Proceedings of the 58th Annual Meeting of the Association
  for Computational Linguistics: System Demonstrations}, pages 187--196,
  Online. Association for Computational Linguistics.

\bibitem[{Houlsby et~al.(2019)Houlsby, Giurgiu, Jastrzebski, Morrone,
  de~Laroussilhe, Gesmundo, Attariyan, and
  Gelly}]{DBLP:conf/icml/HoulsbyGJMLGAG19}
Neil Houlsby, Andrei Giurgiu, Stanislaw Jastrzebski, Bruna Morrone, Quentin
  de~Laroussilhe, Andrea Gesmundo, Mona Attariyan, and Sylvain Gelly. 2019.
\newblock \href {http://proceedings.mlr.press/v97/houlsby19a.html}
  {Parameter-efficient transfer learning for {NLP}}.
\newblock In \emph{Proceedings of the 36th International Conference on Machine
  Learning, {ICML} 2019, 9-15 June 2019, Long Beach, California, {USA}},
  volume~97 of \emph{Proceedings of Machine Learning Research}, pages
  2790--2799. {PMLR}.

\bibitem[{Khurana et~al.(2022)Khurana, Koli, Khatter, and
  Singh}]{khurana2022natural}
Diksha Khurana, Aditya Koli, Kiran Khatter, and Sukhdev Singh. 2022.
\newblock Natural language processing: State of the art, current trends and
  challenges.
\newblock \emph{Multimedia tools and applications}, pages 1--32.

\bibitem[{Lertvittayakumjorn and
  Toni(2021)}]{lertvittayakumjorn-toni-2021-explanation}
Piyawat Lertvittayakumjorn and Francesca Toni. 2021.
\newblock \href {https://doi.org/10.1162/tacl_a_00440} {Explanation-based human
  debugging of {NLP} models: A survey}.
\newblock \emph{Transactions of the Association for Computational Linguistics},
  9:1508--1528.

\bibitem[{Liu et~al.(2021)Liu, Fu, Xiao, Yuan, Chang, Dai, Liu, Ye, and
  Neubig}]{liu-etal-2021-explainaboard}
Pengfei Liu, Jinlan Fu, Yang Xiao, Weizhe Yuan, Shuaichen Chang, Junqi Dai,
  Yixin Liu, Zihuiwen Ye, and Graham Neubig. 2021.
\newblock \href {https://doi.org/10.18653/v1/2021.acl-demo.34}
  {{E}xplaina{B}oard: An explainable leaderboard for {NLP}}.
\newblock In \emph{Proceedings of the 59th Annual Meeting of the Association
  for Computational Linguistics and the 11th International Joint Conference on
  Natural Language Processing: System Demonstrations}, pages 280--289, Online.
  Association for Computational Linguistics.

\bibitem[{Liu et~al.(2019)Liu, Ott, Goyal, Du, Joshi, Chen, Levy, Lewis,
  Zettlemoyer, and Stoyanov}]{liu2019roberta}
Yinhan Liu, Myle Ott, Naman Goyal, Jingfei Du, Mandar Joshi, Danqi Chen, Omer
  Levy, Mike Lewis, Luke Zettlemoyer, and Veselin Stoyanov. 2019.
\newblock Roberta: A robustly optimized bert pretraining approach.
\newblock \emph{arXiv preprint arXiv:1907.11692}.

\bibitem[{Madsen et~al.(2022)Madsen, Reddy, and Chandar}]{madsen2022posthoc}
Andreas Madsen, Siva Reddy, and Sarath Chandar. 2022.
\newblock \href {https://doi.org/10.1145/3546577} {Post-hoc interpretability
  for neural nlp: A survey}.
\newblock \emph{ACM Comput. Surv.}, 55(8).

\bibitem[{Mathew et~al.(2021)Mathew, Saha, Yimam, Biemann, Goyal, and
  Mukherjee}]{DBLP:conf/aaai/MathewSYBG021}
Binny Mathew, Punyajoy Saha, Seid~Muhie Yimam, Chris Biemann, Pawan Goyal, and
  Animesh Mukherjee. 2021.
\newblock \href {https://ojs.aaai.org/index.php/AAAI/article/view/17745}
  {Hatexplain: {A} benchmark dataset for explainable hate speech detection}.
\newblock In \emph{Thirty-Fifth {AAAI} Conference on Artificial Intelligence,
  {AAAI} 2021, Thirty-Third Conference on Innovative Applications of Artificial
  Intelligence, {IAAI} 2021, The Eleventh Symposium on Educational Advances in
  Artificial Intelligence, {EAAI} 2021, Virtual Event, February 2-9, 2021},
  pages 14867--14875. {AAAI} Press.

\bibitem[{Monarch(2021)}]{monarch2021human}
Robert~Munro Monarch. 2021.
\newblock \emph{Human-in-the-Loop Machine Learning: Active learning and
  annotation for human-centered AI}.
\newblock Simon and Schuster.

\bibitem[{Mosca et~al.(2022{\natexlab{a}})Mosca, Agarwal, Rando~Ram{\'\i}rez,
  and Groh}]{mosca-etal-2022-suspicious}
Edoardo Mosca, Shreyash Agarwal, Javier Rando~Ram{\'\i}rez, and Georg Groh.
  2022{\natexlab{a}}.
\newblock \href {https://doi.org/10.18653/v1/2022.acl-long.538} {{``}that is a
  suspicious reaction!{''}: Interpreting logits variation to detect {NLP}
  adversarial attacks}.
\newblock In \emph{Proceedings of the 60th Annual Meeting of the Association
  for Computational Linguistics (Volume 1: Long Papers)}, pages 7806--7816,
  Dublin, Ireland. Association for Computational Linguistics.

\bibitem[{Mosca et~al.(2022{\natexlab{b}})Mosca, Szigeti, Tragianni, Gallagher,
  and Groh}]{mosca-etal-2022-shap}
Edoardo Mosca, Ferenc Szigeti, Stella Tragianni, Daniel Gallagher, and Georg
  Groh. 2022{\natexlab{b}}.
\newblock \href {https://aclanthology.org/2022.coling-1.406} {{SHAP}-based
  explanation methods: A review for {NLP} interpretability}.
\newblock In \emph{Proceedings of the 29th International Conference on
  Computational Linguistics}, pages 4593--4603, Gyeongju, Republic of Korea.
  International Committee on Computational Linguistics.

\bibitem[{Raffel et~al.(2020)Raffel, Shazeer, Roberts, Lee, Narang, Matena,
  Zhou, Li, and Liu}]{raffel2020t5}
Colin Raffel, Noam Shazeer, Adam Roberts, Katherine Lee, Sharan Narang, Michael
  Matena, Yanqi Zhou, Wei Li, and Peter~J. Liu. 2020.
\newblock \href {http://jmlr.org/papers/v21/20-074.html} {Exploring the limits
  of transfer learning with a unified text-to-text transformer}.
\newblock \emph{J. Mach. Learn. Res.}, 21:140:1--140:67.

\bibitem[{Rao and Tetreault(2018)}]{rao-tetreault-2018-dear}
Sudha Rao and Joel Tetreault. 2018.
\newblock \href {https://doi.org/10.18653/v1/N18-1012} {Dear sir or madam, may
  {I} introduce the {GYAFC} dataset: Corpus, benchmarks and metrics for
  formality style transfer}.
\newblock In \emph{Proceedings of the 2018 Conference of the North {A}merican
  Chapter of the Association for Computational Linguistics: Human Language
  Technologies, Volume 1 (Long Papers)}, pages 129--140, New Orleans,
  Louisiana. Association for Computational Linguistics.

\bibitem[{Ray et~al.(2019)Ray, Yao, Kumar, Divakaran, and
  Burachas}]{ray2019can}
Arijit Ray, Yi~Yao, Rakesh Kumar, Ajay Divakaran, and Giedrius Burachas. 2019.
\newblock Can you explain that? lucid explanations help human-ai collaborative
  image retrieval.
\newblock In \emph{Proceedings of the AAAI Conference on Human Computation and
  Crowdsourcing}, volume~7, pages 153--161.

\bibitem[{Ribeiro et~al.(2016)Ribeiro, Singh, and Guestrin}]{ribeiro2016should}
Marco~Tulio Ribeiro, Sameer Singh, and Carlos Guestrin. 2016.
\newblock " why should i trust you?" explaining the predictions of any
  classifier.
\newblock In \emph{Proceedings of the 22nd ACM SIGKDD international conference
  on knowledge discovery and data mining}, pages 1135--1144.

\bibitem[{Scao et~al.(2022{\natexlab{a}})Scao, Fan, Akiki, Pavlick, Ilic,
  Hesslow, Castagn{\'{e}}, Luccioni, Yvon, Gall{\'{e}}, Tow, Rush, Biderman,
  Webson, Ammanamanchi, Wang, Sagot, Muennighoff, del Moral, Ruwase, Bawden,
  Bekman, McMillan{-}Major, Beltagy, Nguyen, Saulnier, Tan, Suarez, Sanh,
  Lauren{\c{c}}on, Jernite, Launay, Mitchell, Raffel, Gokaslan, Simhi, Soroa,
  Aji, Alfassy, Rogers, Nitzav, Xu, Mou, Emezue, Klamm, Leong, van Strien,
  Adelani, and et~al.}]{scao2022bloom}
Teven~Le Scao, Angela Fan, Christopher Akiki, Ellie Pavlick, Suzana Ilic,
  Daniel Hesslow, Roman Castagn{\'{e}}, Alexandra~Sasha Luccioni,
  Fran{\c{c}}ois Yvon, Matthias Gall{\'{e}}, Jonathan Tow, Alexander~M. Rush,
  Stella Biderman, Albert Webson, Pawan~Sasanka Ammanamanchi, Thomas Wang,
  Beno{\^{\i}}t Sagot, Niklas Muennighoff, Albert~Villanova del Moral, Olatunji
  Ruwase, Rachel Bawden, Stas Bekman, Angelina McMillan{-}Major, Iz~Beltagy,
  Huu Nguyen, Lucile Saulnier, Samson Tan, Pedro~Ortiz Suarez, Victor Sanh,
  Hugo Lauren{\c{c}}on, Yacine Jernite, Julien Launay, Margaret Mitchell, Colin
  Raffel, Aaron Gokaslan, Adi Simhi, Aitor Soroa, Alham~Fikri Aji, Amit
  Alfassy, Anna Rogers, Ariel~Kreisberg Nitzav, Canwen Xu, Chenghao Mou, Chris
  Emezue, Christopher Klamm, Colin Leong, Daniel van Strien, David~Ifeoluwa
  Adelani, and et~al. 2022{\natexlab{a}}.
\newblock \href {https://doi.org/10.48550/arXiv.2211.05100} {{BLOOM:} {A}
  176b-parameter open-access multilingual language model}.
\newblock \emph{CoRR}, abs/2211.05100.

\bibitem[{Scao et~al.(2022{\natexlab{b}})Scao, Fan, Akiki, Pavlick, Ilic,
  Hesslow, Castagn{\'{e}}, Luccioni, Yvon, Gall{\'{e}}, Tow, Rush, Biderman,
  Webson, Ammanamanchi, Wang, Sagot, Muennighoff, del Moral, Ruwase, Bawden,
  Bekman, McMillan{-}Major, Beltagy, Nguyen, Saulnier, Tan, Suarez, Sanh,
  Lauren{\c{c}}on, Jernite, Launay, Mitchell, Raffel, Gokaslan, Simhi, Soroa,
  Aji, Alfassy, Rogers, Nitzav, Xu, Mou, Emezue, Klamm, Leong, van Strien,
  Adelani, and et~al.}]{DBLP:journals/corr/abs-2211-05100}
Teven~Le Scao, Angela Fan, Christopher Akiki, Ellie Pavlick, Suzana Ilic,
  Daniel Hesslow, Roman Castagn{\'{e}}, Alexandra~Sasha Luccioni,
  Fran{\c{c}}ois Yvon, Matthias Gall{\'{e}}, Jonathan Tow, Alexander~M. Rush,
  Stella Biderman, Albert Webson, Pawan~Sasanka Ammanamanchi, Thomas Wang,
  Beno{\^{\i}}t Sagot, Niklas Muennighoff, Albert~Villanova del Moral, Olatunji
  Ruwase, Rachel Bawden, Stas Bekman, Angelina McMillan{-}Major, Iz~Beltagy,
  Huu Nguyen, Lucile Saulnier, Samson Tan, Pedro~Ortiz Suarez, Victor Sanh,
  Hugo Lauren{\c{c}}on, Yacine Jernite, Julien Launay, Margaret Mitchell, Colin
  Raffel, Aaron Gokaslan, Adi Simhi, Aitor Soroa, Alham~Fikri Aji, Amit
  Alfassy, Anna Rogers, Ariel~Kreisberg Nitzav, Canwen Xu, Chenghao Mou, Chris
  Emezue, Christopher Klamm, Colin Leong, Daniel van Strien, David~Ifeoluwa
  Adelani, and et~al. 2022{\natexlab{b}}.
\newblock \href {https://doi.org/10.48550/arXiv.2211.05100} {{BLOOM:} {A}
  176b-parameter open-access multilingual language model}.
\newblock \emph{CoRR}, abs/2211.05100.

\bibitem[{Selvaraju et~al.(2019)Selvaraju, Lee, Shen, Jin, Ghosh, Heck, Batra,
  and Parikh}]{selvaraju2019taking}
Ramprasaath~R Selvaraju, Stefan Lee, Yilin Shen, Hongxia Jin, Shalini Ghosh,
  Larry Heck, Dhruv Batra, and Devi Parikh. 2019.
\newblock Taking a hint: Leveraging explanations to make vision and language
  models more grounded.
\newblock In \emph{Proceedings of the IEEE International Conference on Computer
  Vision}, pages 2591--2600.

\bibitem[{Strout et~al.(2019)Strout, Zhang, and
  Mooney}]{strout-etal-2019-human}
Julia Strout, Ye~Zhang, and Raymond Mooney. 2019.
\newblock \href {https://doi.org/10.18653/v1/W19-4807} {Do human rationales
  improve machine explanations?}
\newblock In \emph{Proceedings of the 2019 ACL Workshop BlackboxNLP: Analyzing
  and Interpreting Neural Networks for NLP}, pages 56--62, Florence, Italy.
  Association for Computational Linguistics.

\bibitem[{Tenney et~al.(2020)Tenney, Wexler, Bastings, Bolukbasi, Coenen,
  Gehrmann, Jiang, Pushkarna, Radebaugh, Reif, and
  Yuan}]{tenney-etal-2020-language}
Ian Tenney, James Wexler, Jasmijn Bastings, Tolga Bolukbasi, Andy Coenen,
  Sebastian Gehrmann, Ellen Jiang, Mahima Pushkarna, Carey Radebaugh, Emily
  Reif, and Ann Yuan. 2020.
\newblock \href {https://doi.org/10.18653/v1/2020.emnlp-demos.15} {The language
  interpretability tool: Extensible, interactive visualizations and analysis
  for {NLP} models}.
\newblock In \emph{Proceedings of the 2020 Conference on Empirical Methods in
  Natural Language Processing: System Demonstrations}, pages 107--118, Online.
  Association for Computational Linguistics.

\bibitem[{Wallace et~al.(2019)Wallace, Tuyls, Wang, Subramanian, Gardner, and
  Singh}]{wallace-etal-2019-allennlp}
Eric Wallace, Jens Tuyls, Junlin Wang, Sanjay Subramanian, Matt Gardner, and
  Sameer Singh. 2019.
\newblock \href {https://doi.org/10.18653/v1/D19-3002} {{A}llen{NLP} interpret:
  A framework for explaining predictions of {NLP} models}.
\newblock In \emph{Proceedings of the 2019 Conference on Empirical Methods in
  Natural Language Processing and the 9th International Joint Conference on
  Natural Language Processing (EMNLP-IJCNLP): System Demonstrations}, pages
  7--12, Hong Kong, China. Association for Computational Linguistics.

\bibitem[{Wang et~al.(2021)Wang, Choi, Xu, and Yang}]{wang-etal-2021-putting}
Zijie~J. Wang, Dongjin Choi, Shenyu Xu, and Diyi Yang. 2021.
\newblock \href {https://aclanthology.org/2021.hcinlp-1.8} {Putting humans in
  the natural language processing loop: A survey}.
\newblock In \emph{Proceedings of the First Workshop on Bridging
  Human{--}Computer Interaction and Natural Language Processing}, pages 47--52,
  Online. Association for Computational Linguistics.

\bibitem[{Wolf et~al.(2020)Wolf, Debut, Sanh, Chaumond, Delangue, Moi, Cistac,
  Rault, Louf, Funtowicz, Davison, Shleifer, von Platen, Ma, Jernite, Plu, Xu,
  Le~Scao, Gugger, Drame, Lhoest, and Rush}]{wolf-etal-2020-transformers}
Thomas Wolf, Lysandre Debut, Victor Sanh, Julien Chaumond, Clement Delangue,
  Anthony Moi, Pierric Cistac, Tim Rault, Remi Louf, Morgan Funtowicz, Joe
  Davison, Sam Shleifer, Patrick von Platen, Clara Ma, Yacine Jernite, Julien
  Plu, Canwen Xu, Teven Le~Scao, Sylvain Gugger, Mariama Drame, Quentin Lhoest,
  and Alexander Rush. 2020.
\newblock \href {https://doi.org/10.18653/v1/2020.emnlp-demos.6} {Transformers:
  State-of-the-art natural language processing}.
\newblock In \emph{Proceedings of the 2020 Conference on Empirical Methods in
  Natural Language Processing: System Demonstrations}, pages 38--45, Online.
  Association for Computational Linguistics.

\bibitem[{Yao et~al.(2021)Yao, Chen, Ye, Jin, and Ren}]{yao2021refining}
Huihan Yao, Ying Chen, Qinyuan Ye, Xisen Jin, and Xiang Ren. 2021.
\newblock Refining language models with compositional explanations.
\newblock \emph{Advances in Neural Information Processing Systems},
  34:8954--8967.

\end{thebibliography}
\bibliographystyle{acl_natbib}

\newpage
\onecolumn

\appendix

\section{Backbone API Endpoints} \label{sec:appendix_backbone_api}

\begin{figure}[h!]
    \centering
    \includegraphics[width=\linewidth]{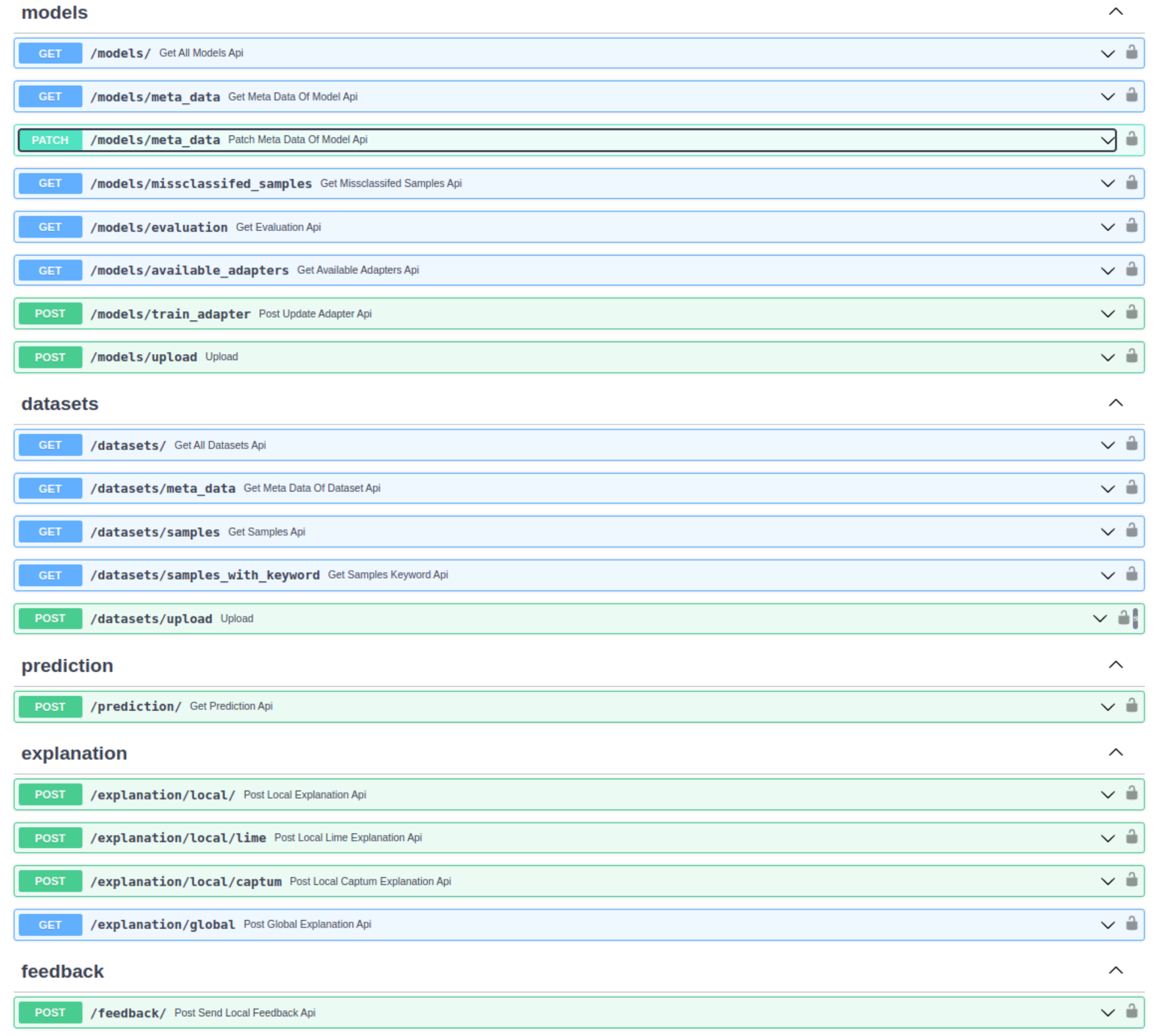}
    \caption{Screenshot of the Swagger UI for our backbone API endpoints.}
    \label{fig:API_overview}
\end{figure}

Figure \ref{fig:API_overview} shows the auto-generated docs for our backbone's REST API, which serves as guidelines to interact with our backbone. Endpoints are divided into functional groups---\emph{models}, \emph{datasets}, \emph{prediction}, \emph{explanation}, and \emph{feedback}). Currently, this page is only accessible within our institution's network for security reasons. 

Developers with direct API access (specifiable on the \emph{configuration page}, see \ref{subsec:user_inteface}) can directly make requests to this high-level interface for additional (larger-scale) experiments. Once again, the API has been built with the Python framework FastAPI\footnote{\href{https://fastapi.tiangolo.com}{https://fastapi.tiangolo.com}}. 

\begin{figure}[th!]
    \centering
    \includegraphics[width=\linewidth]{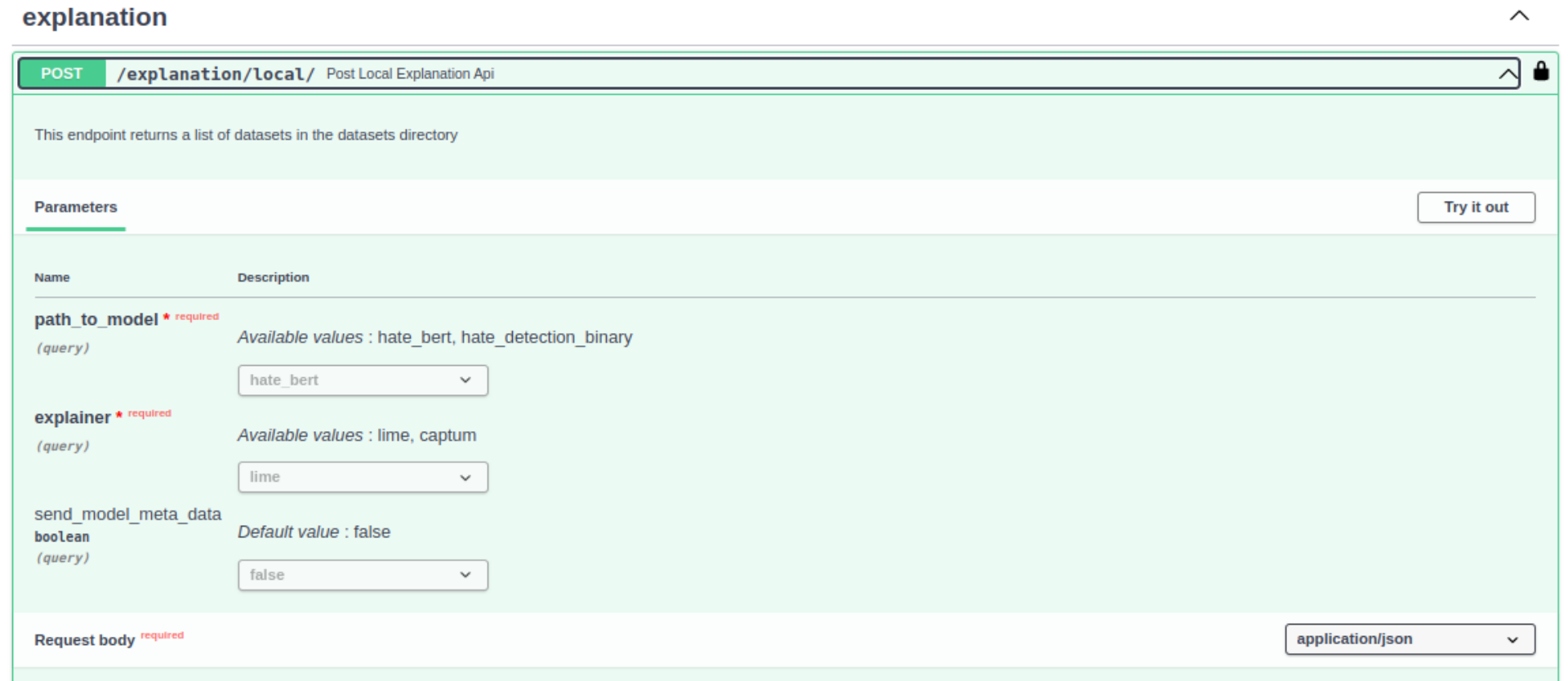}
    \caption{Screenshot of the \emph{explanation} endpoint from our backbone API's Swagger UI.}
    \label{fig:API_endpoint}
\end{figure}

Figure \ref{fig:API_endpoint} shows the documentation for the \emph{explanation} endpoint. Here, we can inspect the details about the endpoint, such as the required parameters---i.e. the path to the model, the explainer to be used (e.g. LIME), and the model's prediction as body request.

\clearpage

\section{User Study} \label{sec:user_study}
We performed a small user study evaluating the usability of our platform with users having different backgrounds. We gathered a total of nine testers, which we organize into three different categories based on their familiarity with the underlying methodologies. Users may be \textit{laymen}, \textit{computer scientists}, or \textit{experts}. For each of these categories, three testers are assigned. Laymen are general users with no exposure to computer science. Computer scientists are those who studied computer science or a related discipline and who work in corresponding fields, such as software development. People who also possess knowledge in the domains of machine learning and NLP are considered experts.

We asked the users to try out to interact with IFAN and then answer several questions about the website’s usability. The questionnaire was of the following structure:
\begin{itemize}
    \item The first question is about the competency of the test person, which is categorized as computer science, natural language processing expert, or layman. \\
    \item Users assign a score between 1 and 5 to the local explanation on the left side of the feedback page. \\
    \item A textual input field for the users to hand in more detailed feedback on the previous rating. \\
    \item Users assign a score between 1 and 5 to the global explanation on the right side of the feedback page. \\
    \item A textual input field for the users to hand in more detailed feedback on the previous rating. \\
    \item Users rate the overall usability of the website with a score between 1 and 5. \\
    \item A textual input field for the users to hand in more detailed feedback on the previous
rating.
\end{itemize}

The average ratings given by users: (i)~Local explanation: 3.88; (ii)~Global explanation: 3.2; (iii)~General usability: 3.33. The problem that the users mentioned were insufficiency of only one form of explanations and feedback, sometimes misleading interface for markup, and waiting time. While some of the problems are already fixed in the current version of the platform, we are planning to consistently improve the usability of the platform. At the same time, we note a positive result---the users with different backgrounds were overall comfortable with IFAN usage and found it useful to increase the controllability of NLP models. We believe this confirms the possibility of IFAN utilization for stakeholders, developers, and annotators in future applications.

\clearpage

\section{Case Study}
In this section, we provide additional information about the experiment conducted under Case Study (Section~\ref{sec:case_study}).

\subsection{Annotated Samples \textit{\textcolor{red}{(Warning: the section contains hate speech only with illustrative purposes. We have no intention to offend the reader and a target racial group.)}}}
\label{sec:app_samples}
Here, we provide the list of that were used for manual annotation (Section \ref{sec:case_study}). These samples were misclassified by hate speech detection model. In Table~\ref{tab:samples_least}, we provide the misclassified samples with the lowest confidence, in Table~\ref{tab:samples_highest} -- with the highest confidence. All the samples and their true labels are taken from the original HateXplain dataset~\cite{DBLP:conf/aaai/MathewSYBG021}. For the label assignment details, refer to the HateXplain dataset description.

\begin{table}[h!]
    \centering
    \begin{tabular}{p{9.3cm}|c|c|c}
    \toprule
        \textbf{Sample} & \textbf{\shortstack{Original \\ Label}} & \textbf{Prediction} & \textbf{\shortstack{Confidence \\ Score}} \\
        \hline
        it not so much about the torah which is pretty innocuous and ignored by many jews despite what is said it really about that forbidden book the talmud or babylonian talmud which as goyim cattle we are not permitted to even read good luck with your research & non-toxic & toxic & 0.56 \\
        \hline
        if memory serves ashkenazi jews and east asians have higher average iqs & non-toxic & toxic & 0.57 \\
        \hline
        speaking of does anyone know which goy group the jews intermarry with in the greatest numbers i was under the impression it was either asians or catholics & toxic & non-toxic & 0.52 \\
        \hline
        <user> the jews faked the moon landing & toxic & non-toxic & 0.52 \\
        \hline
        <user> <user> she will blame it on jews or trump i bet & toxic & non-toxic & 0.53 \\
        \hline
        christianity role is historical protector of the jews it has afforded the tribe the time and space it needed to prepare white genocide teamwhite & toxic & non-toxic & 0.54 \\
        \hline
        the jews are gearing up to help usher in democrats in <number> twitter facebook msm google youtube all upping their censorships of the right pro whites & toxic & non-toxic & 0.51 \\
        \hline
        amusing to think that white people will never be given a referendum on expelling the jews from our nations when you think about things in such terms it becomes abundantly clear that we are expected by our governments to suffer jewish parasitism without the right to defend ourselves & toxic & non-toxic & 0.50 \\
        \hline
        it all hehe haha until we show up at your local favorite hang out spot and wait for you to come out in the van they go one little heeb at a time whatever it takes & toxic & non-toxic & 0.53 \\
        \hline
        <user> <user> jews and the stupid holocaust myth & toxic & non-toxic & 0.52 \\
        \hline
        <user> jews will never be on the blockchain because you can not trust them & toxic & non-toxic & 0.51 \\
        \hline
        <user> <user> it the jews not europeans stop shifting blame on gentiles & toxic & non-toxic & 0.57 \\
    \bottomrule
    \end{tabular}
    \caption{Misclassified with the lowest confidence samples used for manual annotation.}
    \label{tab:samples_least}
\end{table}

\begin{table}[h!]
    \centering
    \begin{tabular}{p{9.3cm}|c|c|c}
    \toprule
        \textbf{Sample} & \textbf{\shortstack{Original \\ Label}} & \textbf{Prediction} & \textbf{\shortstack{Confidence \\ Score}} \\
        \hline
        would rather have terrorism over not being able to boycott an allied country this talking point must be important because he repeats it but i can not make out what on earth it supposed to mean if you do not buy stuff from jews they will fake arabs taking down more nyc landmarks & non-toxic & toxic & 0.72 \\
        \hline
        how long will jews be a majority in israel & non-toxic & toxic & 0.88 \\
        \hline
        what do you mean goes and buys their house not only predates the ghetto it was there before ghetto people came to exist why do you want white people to take a stand for their home but jews to flee theirs & non-toxic & toxic & 0.77 \\
        \hline
        calling me a kike is not an insult & non-toxic & toxic & 0.97 \\
        \hline
        the jews and muslims grew up together and the jews gave birth to christianity catholicism and later on mormonism like a big religion family & non-toxic & toxic & 0.78 \\
        \hline
        greg susan frank rodrick and manny heffley are all jews & non-toxic & toxic & 0.67 \\
        \hline
        it amazing how long <user> talked without giving any answers i learned way more about <user> thoughts and life than sargon i thought only jews could bloviate that hard hm & toxic & non-toxic & 0.95 \\
        \hline
        <user> <user> um akshually star trek were the space commies and star wars jedis are just space jews & toxic & non-toxic & 0.87 \\
        \hline
        <user> apparently not so obvious to some esp the homosexual jewish media that wants women to look like boys and people who think that is traditional rather than feminist & toxic & non-toxic & 0.84 \\
        \hline
        sorry i was referring to the holocaust story i believe it to be a lie & toxic & non-toxic & 0.89 \\
        \hline
        i swear college is a hot house full of judaised professors with white youth being indoctrinated into this nonsense or have live there days out in fear of being ousted as racists and proud members of white race & toxic & non-toxic & 0.90 \\
        \hline
        a few months ago i said that anti semitism doesnt exists naming a group of people is a branch of social sciences called sociology well jews are at war with social sciences now you simply cant name them & toxic & non-toxic & 0.74 \\
    \bottomrule
    \end{tabular}
    \caption{Misclassified with the highest confidence samples used for manual annotation.}
    \label{tab:samples_highest}
\end{table}

\clearpage

\subsection{Feedback Mechanisms Comparison}
\label{sec:app_eddback_comparison}
\begin{figure}[th!]
    \centering
    \begin{subfigure}{0.46\textwidth}
        \includegraphics[width=0.96\textwidth]{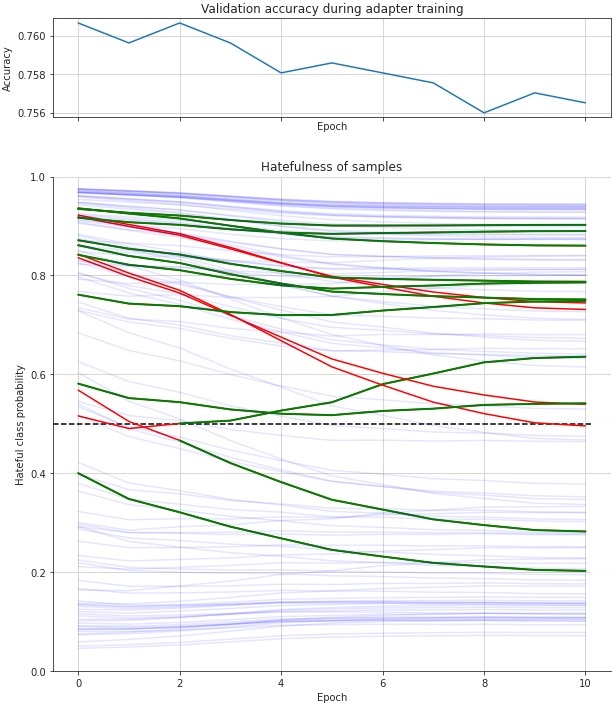}
        \caption{Training without feedback balancing.}
    \end{subfigure}
    \begin{subfigure}{0.46\textwidth}
        \includegraphics[width=0.96\textwidth]{img/least_conf_balanced.jpg}
        \caption{Training with feedback balancing.}
    \end{subfigure}
    \caption{The comparison of training procedure with and without feedback balancing. Here, the results of local feedback on the least confident misclassified samples from the Jewish target group are shown. We can observe that training with a balanced dataset runs more stable without significant influence on the overall model's domain knowledge.}
    \label{fig:app_results_local}
\end{figure}

\begin{figure}[th!]
    \centering
    \begin{subfigure}{0.46\textwidth}
        \includegraphics[width=0.96\textwidth]{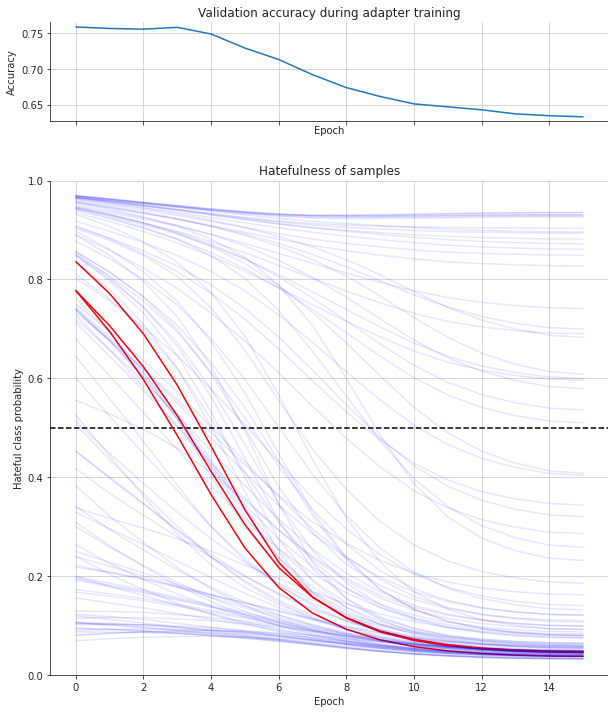}
        \caption{Training without feedback balancing.}
    \end{subfigure}
    \begin{subfigure}{0.46\textwidth}
        \includegraphics[width=0.96\textwidth]{img/global_injection_cut.png}
        \caption{Training with feedback balancing.}
    \end{subfigure}
    \caption{The comparison of training procedure with and without feedback balancing. Here, the results of local feedback on misclassified samples with ``jewish'' keywords are shown. We can observe that training with rebalanced dataset runs more stable without significant influence on the overall model's domain knowledge.}
    \label{fig:app_results_global}
\end{figure}

In Section~\ref{sec:case_study}, we report the results of the model trained on feedback in two setups: (i)~without balancing and (ii)~with balancing via using original samples from HateXplain dataset. The comparison between these two setups is visualized in Figures~\ref{fig:app_results_local} and \ref{fig:app_results_global}. We tested our approach on the local feedback on the Jewish target group samples as well as samples containing the ``Jewish'' keyword. For both setups, with balancing, the training procedure runs more stable. The model's performance on other samples from HateXplain dataset changes slightly and the adjustment of its behavior on the marked-up samples proceeds more rapidly.

\clearpage

\newpage

\section{BLOOM for Hate Speech Classification and Human Feedback Incorporation}
\label{sec:app_bloom}
\begin{figure}[h!]
    \centering
    \includegraphics[width=\textwidth]{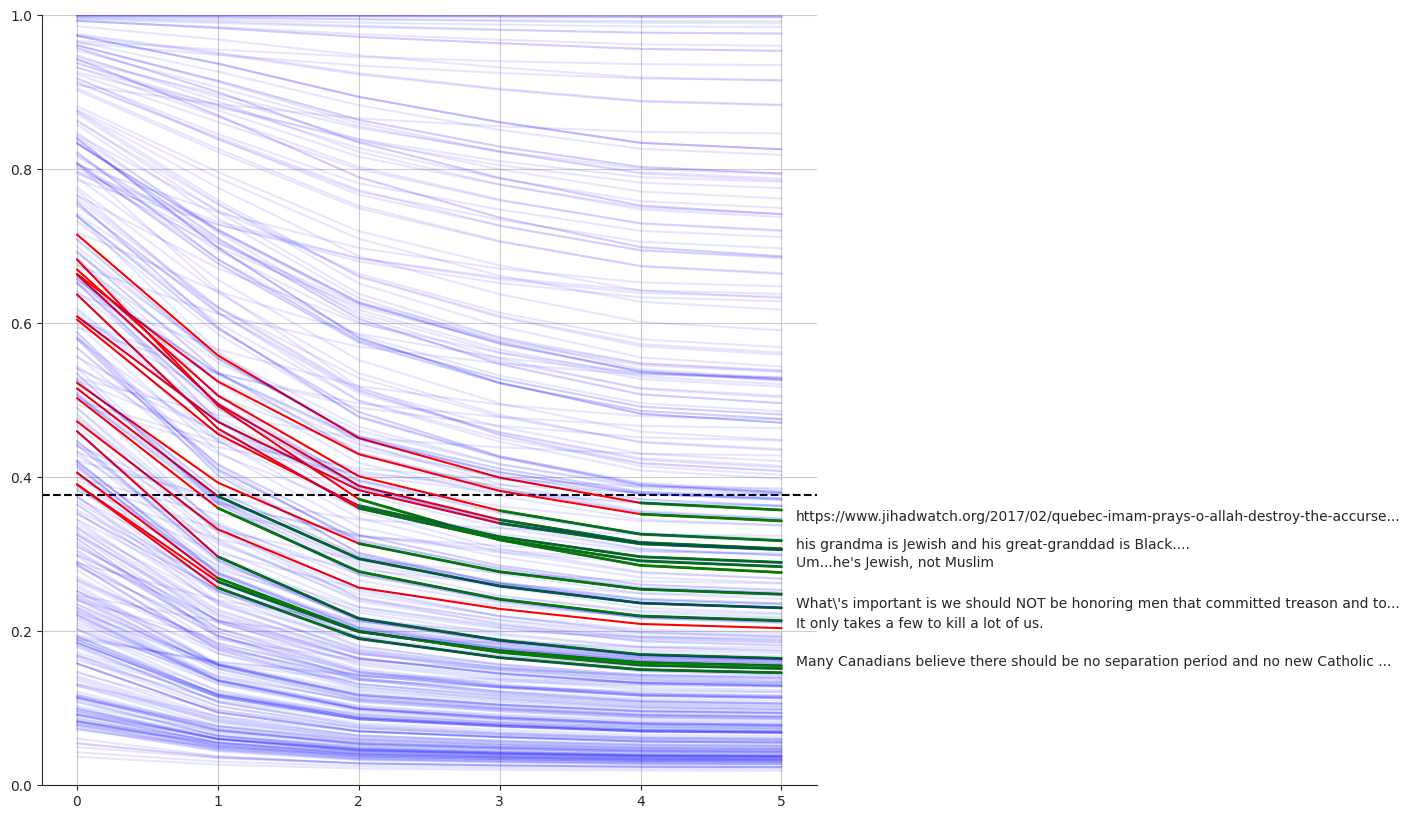}
    \caption{The tracking procedure of the misclassified samples label fix during feedback incorporation.}
    \label{fig:bloom_feedback}
\end{figure}

In addition to the results with BERT for hate speech classification (Section~\ref{sec:case_study}), we provide the results of experiments with a BLOOM\footnote{\href{https://huggingface.co/bigscience/bloom-560m}{https://huggingface.co/bigscience/bloom-560m}} model~\cite{DBLP:journals/corr/abs-2211-05100} (Table~\ref{tab:bloom_study_results}). Once again, the fine-tuning of adapter layer with human feedback improves model performance within Jewish target group. In the case of BLOOM, the overall F1 score is also improved, confirming the utility of human feedback usage and the possibility of it being extended to other models. Also, from Figure~\ref{fig:bloom_feedback}, we can see how the label of the misclassified samples (i.e. \textit{"Um ... he's jewish, not Muslim"}) is changing as the model is fine-tuned with the provided feedback.

\begin{table}[h!]
    \centering
    \footnotesize
    \begin{tabular}{p{3.5cm}|c|c|c|c}
    \toprule
        \textbf{Model} & \textbf{Pr} & \textbf{Re} & \textbf{F1} & \textbf{F1$_{J}$}\\
        \hline
        BLOOM (baseline) & 0.40 & \textbf{0.70} & 0.48 & 0.48 \\
        BLOOM+Feedback & \textbf{0.49} & 0.58 & \textbf{0.51} & \textbf{0.53} \\
    \bottomrule
    \end{tabular}
    \caption{The results of the LLMs inference for hate speech classification. F1$_{J}$ states for the F1 score on the Jewish target group.}
    \label{tab:bloom_study_results}
\end{table}

\section{Large Language Models for Hate Speech Classification and Explanation}
\label{sec:app_llm}
In addition to the classification via linear layers on top of transformer blocks, we test the possibility of large language models to perform text classification and explain their prediction. Potentially, this approach allows further work to extend the framework to any arbitrary NLP tasks. 

Currently, the models available on IFAN for testing are: vicgalle/gpt2-alpaca\footnote{\href{https://huggingface.co/vicgalle/gpt2-alpaca}{https://huggingface.co/vicgalle/gpt2-alpaca}}, chavinlo/alpaca-native\footnote{\href{https://huggingface.co/chavinlo/alpaca-native}{https://huggingface.co/chavinlo/alpaca-native}}, and chavinlo/gpt4-x-alpaca.\footnote{\href{https://huggingface.co/chavinlo/gpt4-x-alpaca}{https://huggingface.co/chavinlo/gpt4-x-alpaca}} Their results on hate speech classification task based on HateXapline dataset~\cite{DBLP:conf/aaai/MathewSYBG021} are presented in Table~\ref{tab:llms_study_results}.

For each model, we optimize the prompts to perform hate speech detection:
\begin{description}
    \item[vicgalle/gpt2-alpaca] \textit{Instruction: perform the text-classification task on the text using the following labels: 0 or 1. Text: would be hilarious if the un building was turned into homeless shelter or hell why not a refugee center The text is 0 Text: {[}DATASET SAMLPE{]}  The text is 1 Text: {[}INPUT TEXT{]} The text is }

    \item[chavinlo/alpaca-native] \textit{This is an overall text-classification. For the given TEXT, CLASSIFY the LABEL of that.The LABEL can be 0 or 1. Just output the LABEL. NOTHING ELSE! TEXT: {[}5 DATASET SAMLPES{]} LABEL: 0 TEXT: {[}5 DATASET SAMLPES{]} LABEL: 1 TEXT: {[}INPUT TEXT{]}.\\ LABEL:}

    \item[chavinlo/gpt4-x-alpaca] \textit{INSTRUCTION: Given the following DATASET DESCRIPTION, EXTRACT the TASK of it, and PERFORM the TASK on the INPUT TEXT. The FINAL LABEL could be 0 or 1. Just output the FINAL LABEL. NOTHING ELSE! DATASET DESCRIPTION: Hatexplain is the first benchmark hate speech dataset covering multiple aspects of the issue. Each post in the dataset is annotated from three different perspectives: the basic, commonly used 3-class classification (i.e., hate, offensive or normal), the target community (i.e., the community that has been the victim of hate speech/offensive speech in the post), and the rationales, i.e., the portions of the post on which their labelling decision (as hate, offensive or normal) is based. CONTEXT: TEXT: {[} DATASET SAMPLE {]} FINAL LABEL: 0 TEXT: {[} DATASET SAMPLE {]} FINAL LABEL: 1 INPUT TEXT: TEXT: {[}INPUT TEXT{]}. FINAL LABEL: }
\end{description}

Despite the specific differences in the various prompts, they all follow the core idea of i)~mention the main task which can be extracted from the dataset metadata; ii)~provide the general information about labels; iii)~provide some examples for each label from the dataset. Potentially, this prompt design can be used for any classification task. 

\begin{table}[h!]
    \centering
    \footnotesize
    \begin{tabular}{p{3.5cm}|c|c|c}
    \toprule
        \textbf{Model} & \textbf{Pr} & \textbf{Re} & \textbf{F1} \\
        \hline
         vicgalle/gpt2-alpaca &  0.63 & 0.56 & 0.59\\
         chavinlo/alpaca-native & 0.60 & 0.51 & 0.55\\
         chavinlo/gpt4-x-alpaca & \textbf{0.66} & \textbf{0.63} & \textbf{0.64}\\
    \bottomrule
    \end{tabular}
    \caption{The results of the LLMs inference for hate speech classification.}
    \label{tab:llms_study_results}
\end{table}


\section{Supplementary Video Demo}
A supplementary video showcase can be found on Youtube\footnote{\href{https://youtu.be/EzC6HI3JwaQ}{https://youtu.be/EzC6HI3JwaQ}}. For completeness, we also point at an additional version of such demo, as well on Youtube\footnote{\href{https://www.youtube.com/watch?v=BzzoQzTsrLo}{https://www.youtube.com/watch?v=BzzoQzTsrLo}}, dating back to March 2023.

\end{document}